\documentclass[10pt,twocolumn,letterpaper]{article}

\usepackage{iccv}
\usepackage{times}
\usepackage{epsfig}
\usepackage{graphicx}
\usepackage{amsmath}
\usepackage{amssymb}

\usepackage{booktabs}
\usepackage{xcolor,colortbl}
\usepackage{multirow}

\usepackage[pagebackref=true,breaklinks=true,colorlinks,bookmarks=false]{hyperref}

\newcommand{\image}{I}
\newcommand{\imageset}{\mathcal{I}}
\newcommand{\visconcept}{v}
\newcommand{\visconceptset}{\mathcal{V}}
\newcommand{\semconcept}{w}
\newcommand{\semconceptset}{\mathcal{W}}
\newcommand{\conceptset}{\Omega}
\newcommand{\pairset}{\mathcal{S}}
\newcommand{\nat}{\mathbb{N}}
\newcommand{\real}{\mathbb{R}}
\newcommand{\sentence}{s}
\newcommand{\textdomain}{\mathcal{L}}
\newcommand{\encoder}{f}
\newcommand{\decoder}{g}
\newcommand{\translator}{h}
\newcommand{\discriminator}{D}
\newcommand{\textfeature}{\phi}
\newcommand{\textfeatureset}{\Phi}
\newcommand{\imgfeature}{\psi}
\newcommand{\imgfeatureset}{\Psi}
\newcommand{\graph}{\mathcal{G}}
\newcommand{\edge}{P}

\definecolor{Gray}{gray}{0.9}

\iccvfinalcopy %

\def\iccvPaperID{242} %

\ificcvfinal\fi
\begin{document}

\title{Towards Unsupervised Image Captioning with Shared Multimodal Embeddings}

\author{Iro Laina\\
Technische Universit\"at M\"unchen\\
{\tt\small iro.laina@tum.de}
\and
Christian Rupprecht\\
University of Oxford\\
{\tt\small chrisr@robots.ox.ac.uk}
\and
Nassir Navab\\
Technische Universit\"at M\"unchen\\
{\tt\small nassir.navab@tum.de}
}

\maketitle

\begin{abstract}
   Understanding images without explicit supervision has become an important problem in computer vision. 
   In this paper, we address image captioning by generating language descriptions of scenes without learning from annotated pairs of images and their captions.
   The core component of our approach is a shared latent space %
   that is structured by visual concepts. 
   In this space, the two modalities should be indistinguishable.
   A language model is first trained to encode sentences into semantically structured embeddings.
   Image features that are translated into this embedding space can be decoded into descriptions through the same language model, similarly to sentence embeddings. 
   This translation is learned from weakly paired images and text using a loss robust to noisy assignments and a conditional adversarial component.
   Our approach allows to exploit large text corpora outside the annotated distributions of image/caption data.  
   Our experiments show that the proposed domain alignment learns a semantically meaningful representation which outperforms previous work.
   
\end{abstract}

\section{Introduction}

\begin{figure}[t]
    \centering
    \includegraphics[clip, trim=0.5cm 2.5cm 0.5cm 0.2cm, width=0.99\columnwidth]{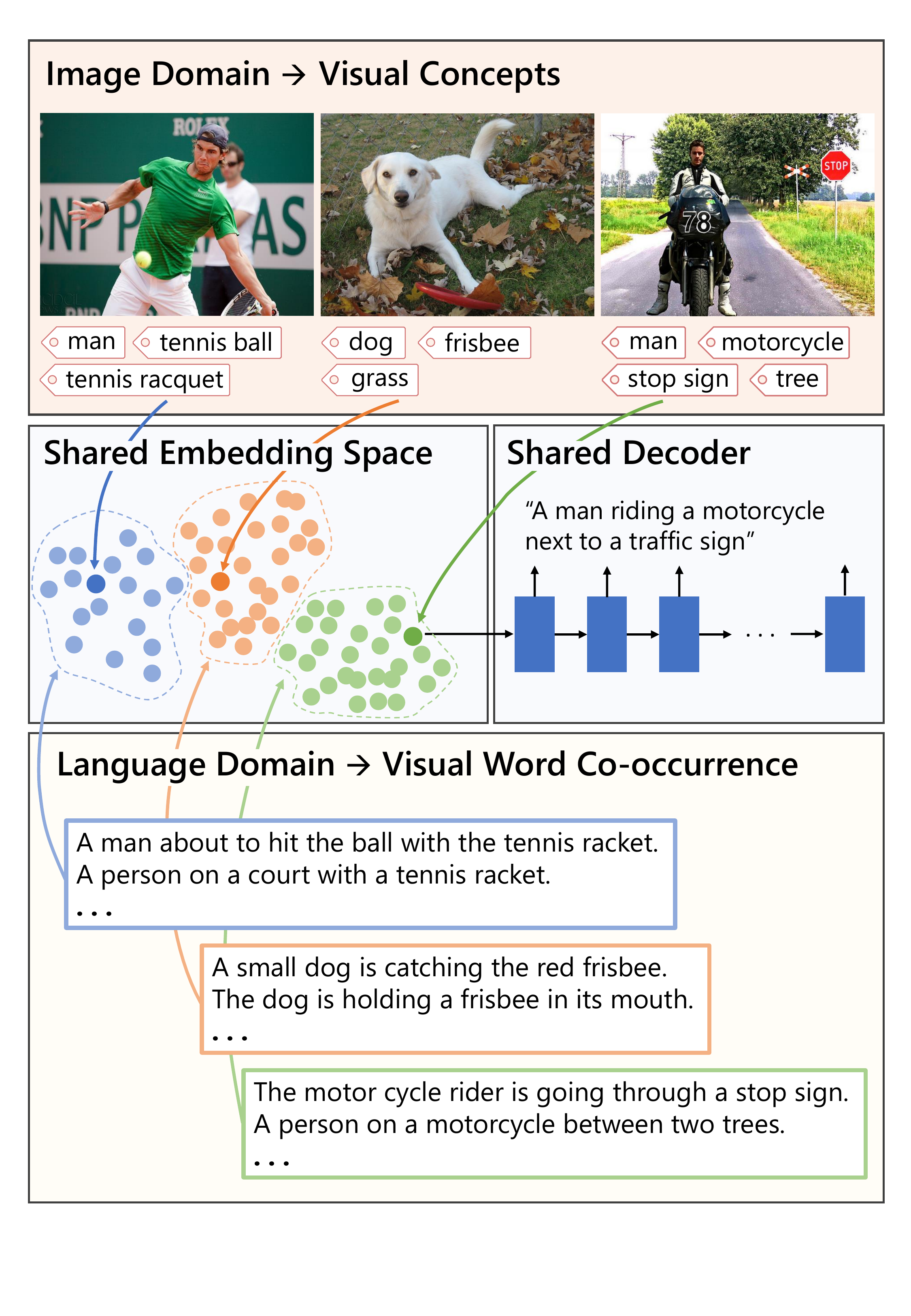}
    \caption{\textbf{Method overview.} Our model learns a joint embedding space of language and image features that is structured by visual concepts and their co-occurrence. Images and text come from disjoint sources. 
    During inference, model embeds images into the shared space from which a caption can be decoded.}
    \label{fig:teaser}
\end{figure}

Generating natural language descriptions for images has gained attention as it aims to teach machines how humans see, understand and talk about the world. 
Assisting visually impaired people~\cite{gurari2018vizwiz,wu2017automatic} and human-robot interaction~\cite{das2017visual,ling2017teaching} are some examples of the importance of image captioning. 
Even though it is straightforward for humans to describe the contents of a scene, machine generation of image descriptions is a challenging problem that requires 
compositional perception of images 
translated into semantically and grammatically correct sentences.

Traditionally, image captioning has been carried out using full supervision in the form of image-caption pairs, given by human annotators. 
Crowd-sourcing captions
is a cumbersome task that requires extensive quality control and further manual cleaning. 
Since annotators are often paid per image, the captions tend to be short and repetitive.
In addition, current captioning benchmarks~\cite{lin2014microsoft,plummer2015flickr30k} consist of a limited number of object categories and are focused on performance under imperfect evaluation metrics. 
Thus, methods developed on such datasets might not be easily adopted in the wild. 
Nevertheless, great efforts have been made to extend captioning to out-of-domain data~\cite{anderson2018partially,chen2017show,zhao2017dual} or different styles beyond mere factual descriptions~\cite{Guo_2019_CVPR,shuster2019engaging}.

In this work we explore \textit{unsupervised} captioning, where image and language sources are independent. 
The unsupervised setting can benefit from an almost unlimited amount of unlabeled or weakly labeled images as well as readily available large text corpora, without the need of bias-prone and costly human annotations.
Although %
significant progress has been achieved in other unsupervised tasks \cite{donahue2016adversarial, lample2017unsupervised,shu2018deforming, zhu2017unpaired}, unsupervised generation of image descriptions remains mostly unexplored. 

The building blocks of our method are a language model and the translation of the image to the language domain.  
On the language side, we first learn a semantically structured embedding space, \ie sentences describing similar visual concepts (\eg \textit{woman} and \textit{person}) and similar context are encoded with similar embeddings. 
We then perform a weakly supervised domain alignment between image features and the learned text embeddings leveraging visual concepts in the image. 
This alignment allows to exploit co-occurrence statistics of visual concepts between sentences and images.
For example, the words \textit{boat} and \textit{water} might often appear together in the language domain, similar to the fact that most images that contain a boat also contain water. 

When language and images come from different sources, some \emph{weak} supervisory signal is needed to align the manifold of visual concepts to the textual domain. 
Similar to previous work \cite{feng2018unsupervised}, we use a pre-trained object detector to generate an initial noisy alignment between the text source and visual entities that can be detected in the image.

We show that we can indeed learn to predict meaningful captions for images that extend beyond the limited capabilities of the object detector. Due to visual concept co-occurrence, the model learns to produce text descriptions including concepts that are not necessarily contained in the object detector's fixed set of labels (\eg \textit{beach}). 
This shows that the alignment is meaningful and the statistics of both domains help to discover more visual concepts. Quantitatively, our unsupervised approach nearly matches the performance of some early supervised methods and outperforms previous unsupervised methods. 
Finally, our approach makes it possible to leverage various language sources, for instance from a different language or with a particular style ---poetic (Shakespeare), funny, story-telling--- that cannot be easily obtained by crowdsourcing.       

\section{Related Work}

\paragraph{Fully supervised.}
Pioneering work in neural-based image captioning~\cite{karpathy2015deep,vinyals2015show} established the commonly used framework of a Convolutional Neural Network (CNN) image encoder, followed by a Recurrent Neural Network (RNN) language decoder. 
There has been significant progress improving over the standard CNN-RNN approach. Xu~\etal~\cite{xu2015show} introduced the concept of attention to image captioning and, subsequently, several methods
focused on attention mechanisms to visualize the grounding of words on image context and effectively guide the generation process~\cite{anderson2018bottom,lu2017knowing,yang2016review,you2016image}.   
Noteworthy efforts also include 
generating video descriptions~\cite{donahue2015long} 
or dense captions on image regions~\cite{johnson2016densecap}, 
exploiting additional information such as attributes~\cite{yao2017boosting}
or visual relationships~\cite{yao2018exploring} and 
optimizing evaluation metrics~\cite{liu2017improved,rennie2017self}. 
Other methods focus on generating diverse and natural captions with adversarial models~\cite{dai2017towards,li2018generating,shetty2017speaking,wang2017diverse}, moving beyond just factual descriptions~\cite{gan2017stylenet,shuster2019engaging} or
addressing gender bias~\cite{anne2018women}. 

\vspace{-1em}
\paragraph{Novel object captioning.}
Recent approaches have also explored the task of
novel object captioning to exploit large-scale visual object recognition from readily available datasets, such as ImageNet~\cite{russakovsky2015imagenet}. 
Their goal is to address the limitations of conventional models in integrating new entities into image descriptions without explicit training pairs.  
In \cite{mao2015learning} the problem is addressed by learning from few labeled pairs for novel categories. 
Copying mechanisms are employed in ~\cite{anne2016deep,yao2017incorporating} to transfer knowledge from the paired data to out-of-domain objects, while \cite{venugopalan2017captioning} jointly exploits semantic information from independent images and text sources. 
Another approach is to produce sentence templates and fill in the slots with detected concepts~\cite{lu2018neural}. 
Instead of training the model to handle new concepts, \cite{Anderson2017GuidedOV} proposes to constrain beam search evaluation on target words. %

\vspace{-1em}
\paragraph{Partial supervision.}
Recent work has further advanced the field towards generating image descriptions under more challenging settings, for example unpaired or unsupervised. 

Chen~\etal~\cite{chen2017show} address cross-domain captioning, 
where the source domain consists of image-caption pairs and the goal is to leverage unpaired data from a target domain through a critic. 
In \cite{zhao2017dual}, the cross-domain problem is addressed with a cycle objective. 
Similarly, unpaired data can be used to generate stylized descriptions ~\cite{Guo_2019_CVPR,mathews2018semstyle}.
Anderson~\etal~\cite{anderson2018partially} propose a method to complete partial sequence data, \eg a sequence of detected visual concepts, without the need for paired image-caption datasets. 
Gu~\etal~\cite{gu2018unpaired} address unpaired image captioning from a different perspective, using an intermediary language where paired data is available, and then translating the captioner to the target language using parallel corpora. 
However, the goal of these methods is different to ours, as they typically align a target domain that contains limited paired or unpaired data with a source domain. 
A generic image captioner is first built from \emph{full} supervision in the source domain and then adapted to a different language domain or novel object categories. 

Most closely related to our work is \cite{feng2018unsupervised} which does not require any image-sentence pairs.
In this case, it is optimal to use a language domain which is rich in visual concepts. 
Therefore, their (and our) goal is to exploit image and language sources that are disjoint yet compatible, instead of aligning different language sources 
as in cross-domain approaches. 
Supervision comes in only through image recognition models, which are used to detect objects in the image. 

\vspace{-1em}
\paragraph{Multimodal embeddings.}
A key component of our approach is the alignment of latent representations from two independent modalities. 
In unsupervised machine translation, although unimodal, \cite{lample2017unsupervised,lample2018phrase} create a shared latent space (interlingua) for both source and target languages.
Kiros~\etal~\cite{kiros2014unifying} pose captioning as a translation problem and learn a multimodal embedding space 
that also allows them to perform vector arithmetics. 
Similarly, joint embedding spaces have been used in \cite{faghri2017vse++} for cross-modality retrieval and in \cite{pan2016jointly} for video captioning. 
Finally, Fang~\etal~\cite{fang2015captions} predict visual words from images to produce caption candidates and use the similarity between images and sentences in a joint space to rank the captions.

\section{Methods}

\begin{figure*}[t]
    \centering
    \includegraphics[clip, trim=0cm 5cm 2cm 0.1cm, width=0.95\textwidth]{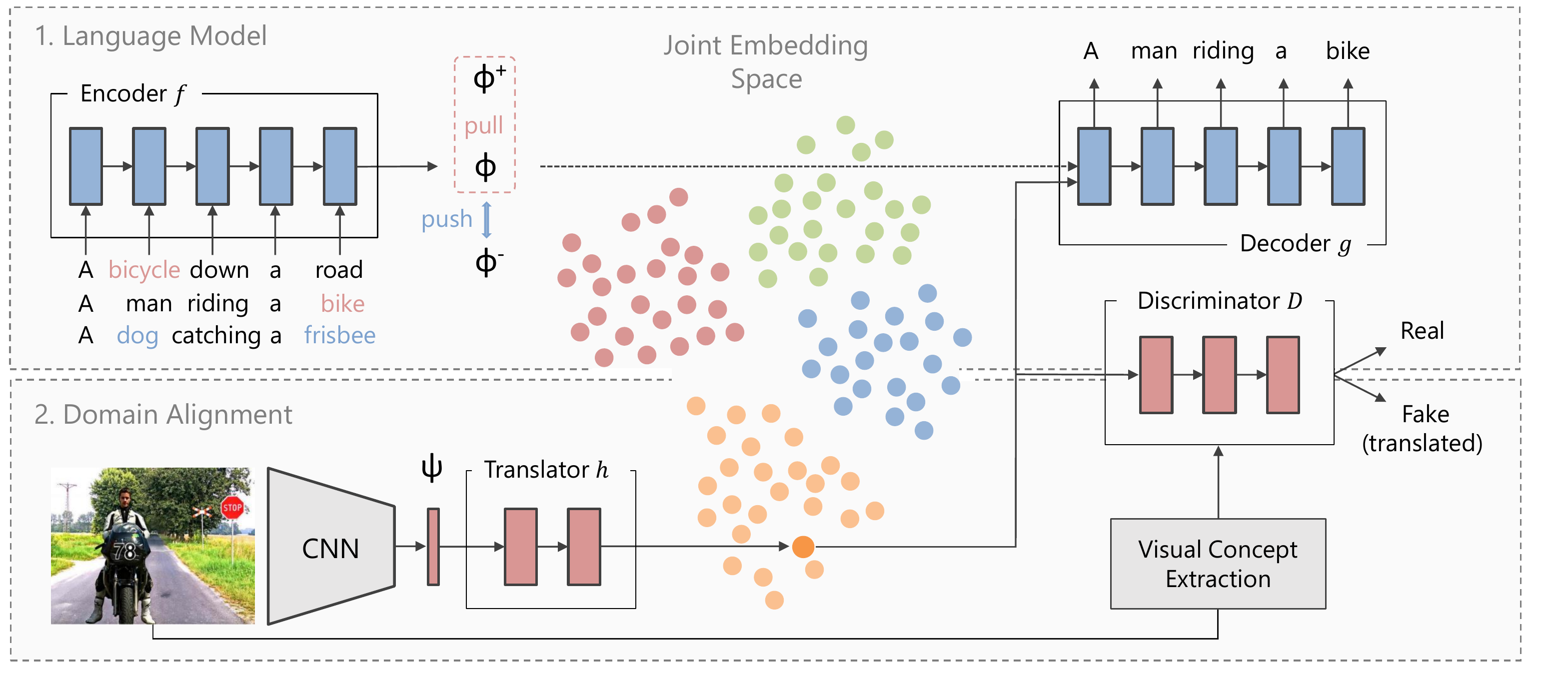}
    \caption{\textbf{Unsupervised image captioning architecture.} We first learn a language model with a triplet loss formulation that structures the embedding $\textfeature$ using visual concepts from the sentences. We then learn a mapping from images to the embedding space using a robust alignment scheme and adversarial training in feature space.}
    \label{fig:method}
    \vspace{-0.2em}
\end{figure*}

An overview of our method is shown in Figure~\ref{fig:method}. 
The proposed approach consists of two components, a language model and a domain alignment model between images and text. 
The language model %
independently encodes samples from the language domain into a semantic-aware representation. 
The goal of the domain alignment is to translate image respresentations into the embedding space learned by the language model and decode these embeddings into meaningful image descriptions.
In absence of paired image-caption data this is a challenging task. 

We consider a visual domain $\imageset$ and an image $\image_i \in \imageset$, represented by the set of visual entities that it encloses:
\begin{equation}
    \visconceptset_i = \{ \visconcept_k \mid k\in \nat,\; 1 \leq k \leq N_i\},  %
    \label{eq:img_synsets}
\end{equation}
where $i$ iterates over the total number of image samples and $N_i$ is the total number of visual concepts in image $i$. 

Similarly, in the language domain $\textdomain$, a text sequence $\sentence_j \in \textdomain$ 
can be described by a bag of words
\begin{equation}
    \semconceptset_j = \{ \semconcept_k \mid k\in \nat,\; 1 \leq k \leq M_j\}, %
    \label{eq:txt_synsets}
\end{equation}
where $j$ enumerates sequences of length $M_j$.  

For the purpose of this work, we assume that the image and language domains are not entirely disjoint. For example, it would seem unreasonable to attempt describing natural images based on text corpora of economics. 
Thus, we assume %
a universal set of concepts $\conceptset = \visconceptset \,\cap \, \semconceptset$ that language and images have in common. %
We refer to joint concepts, such as \textit{person}, as visual concepts.

\subsection{Language Model}
\label{sub:language_model}

To create a basis for domain alignment, our first step is to create a meaningful textual domain.
We learn an unsupervised sentence embedding by training a language model on the text corpus,
following a standard sequence-to-sequence approach with maximum likelihood estimation~\cite{sutskever2014sequence}. 
The encoder $\encoder$ embeds an input sentence $\sentence$ into a $d$-dimensional latent representation which is reconstructed back into the same sentence by a decoder $\decoder$:
\begin{equation}
\label{eq:lm}
\encoder(\sentence) \,=\, \textfeature , \quad
\decoder(\textfeature) \,=\, \tilde{\sentence} , \quad \textfeature \in \textfeatureset \subseteq \real^d.
\end{equation}
RNNs are the most common choice for $f$ and $g$. 
Typically, language models of this structure are trained by minimizing the negative log-likelihood between $\sentence$ and $\tilde{\sentence}$ per word.

A model without any constraints on the latent space would learn a grammatical and syntactic embedding. Instead, we are primarily interested in creating a representation that encodes visual semantics.
This means that we have to encourage the model to learn a manifold structured by visual concepts.
As we show later, our representation encodes strong semantic properties in the sense that sentences with similar contents have a low distance in the embedding space. 
Since our goal is image captioning, our notion of \textit{similar sentence contents} stems from visual concepts ---words in a sentence that have visual grounding--- and their co-occurrence statistics. 
We impose a visual concept-based structure on the manifold of $\textfeature$
with a triplet loss, defined as 
\begin{equation}
    L_t(\,\textfeature, \textfeature^+, \textfeature^-) = \max(0,\; \|\textfeature - \textfeature^+\|_2^2 - \|\textfeature - \textfeature^-\|_2^2 \,+\, m)
    \label{eq:triplet_loss}
    \vspace{-0.2em}
\end{equation}
that operates on triplets of embeddings $\textfeature$. The loss is minimized when the distance from an anchor embedding $\textfeature$ to a \textit{positive pair} $\textfeature^+$ is smaller than the distance to a \textit{negative pair} $\textfeature^-$ by at least a margin $m \in \real^+$. 

The positive and negative pairs can be defined based on the visual concepts that exist in the sentences. For a given sentence $\sentence_j$ we define the set of negative pairs $\pairset_j^-$ as the set of sentences that do not have any concepts in common 
\begin{equation}
    \pairset_j^- = \{ \sentence_k \mid k \in \nat ,\, \semconceptset_k \cap \semconceptset_j = \emptyset\}\,.
\end{equation}
Analogously, we define the set of positive pairs $\pairset_j^+$ as the set of sentences that have at least \textit{two} concepts in common 
\begin{equation}
    \pairset_j^+ = \{ \sentence_k \mid k \in \nat ,\, k \ne j,\, |\semconceptset_k \cap \semconceptset_j| \ge 2 \}\,.
\end{equation}

We ignore sentence pairs that only have one overlapping concept to reduce bad alignments. For example, since many language datasets are human-centered, every sentence involving a person would be a positive pair to each other regardless of the context. 
The language model's total loss is
\begin{equation}
    L_{\operatorname{LM}}(\sentence_j) =
    L_{\operatorname{CE}}(\,\decoder(\textfeature),\, \sentence_j\,) \,+\,
    \lambda_{t}\,L_t(\textfeature_j, \textfeature^+_j, \textfeature^-_j)   .
\end{equation}

During training, a positive sentence $\sentence^+ \in \pairset_j^+$ is sampled from a multinomial distribution with probability proportional to the number of overlapping concepts. This favors positive pairs of sentences with many similar concepts. We sample a negative sentence $\sentence^-$
uniformly from $\pairset_j^-$. 

The triplet loss imposes a visually aware structure on the embedding space. 
Sentences with similar visual contents are encouraged to be close to each other, while sentences with different context will be pushed apart. 
This external emphasis on structure is important, since unconstrained language models are more likely to group sentences with similar words and grammar. 
Intuitively, generating image descriptions relies on visual content and thus the structured embedding space is presumably a more meaningful basis for the task at hand.   
A comparison between the visually constrained and unconstrained embedding space can be found in the supplementary material. 

\subsection{Joint Image and Language Domain}
We have learned an encoder that projects text into a structured embedding. 
The next step is to project image features into the same embedding space so that they can be similarly decoded into sentences by the decoder.
To do this, we need an initial alignment between the independent image and text sources for which we rely on the visual concepts they have in common. 
We build a bipartite graph $\graph(\textdomain, \imageset, \edge)$ with images $\image_i$ and sentences $\sentence_j$ as nodes. The edges $\edge_{i,j}$ represent weak assignments between $\image_i$ and $\sentence_j$, weighted by the number of overlapping concepts 
\begin{equation}
    \edge_{i,j} \,=\, |\visconceptset_i \cap \semconceptset_j| .
    \label{eq:ass_graph}
\end{equation}

During training, for $\image_i$ we sample $\sentence_j$ with probability
\begin{equation}
\label{eq:prob}
    p(\sentence_j \mid \image_i) \,=\, \edge_{i,j} \; \Big( \sum_k \edge_{i,k} \Big)^{-1} .
\end{equation}

For sentence-image pairs without overlap $p(\sentence_j \mid \image_i) = 0$ and they are excluded from training. Highly visually correlated pairs will be sampled with higher probability. 
At this point, we have created a stochastic training set, which we could use to train a standard captioning model by sampling an image-caption pair at each iteration. 
Training this model with teacher forcing alone, collapses to certain caption-modes describing sets of images.

Visual concepts can be extracted from the images using any pretrained image recognition method. However, this would often result in only a limited number of categories. To lexically enrich the search space for matching sentences, we also query hyponyms of the predicted visual concepts $\visconceptset_i$, \ie words among the text source concepts $\semconceptset_i$ that have a kind-of relationship with the predicted concepts (for example, man to person, puppy to dog).

\subsection{Learning the Semantic Alignment}
The initial alignment allows us to learn a mapping from images to text. We extract image features $\imgfeature_i$ from $\image_i$ using a standard pretrained CNN. The task is now to translate between the image feature domain $\imgfeature_i \in \imgfeatureset$ to the visually structured text domain $\textfeature_j \in \textfeatureset$.
The stochastic alignment graph $\graph$ is expected to be very noisy and full of imprecise correspondences. 
We thus propose a robust training scheme to exploit the underlying co-occurence information while ignoring problematic matches. 
We learn the translation function $\translator : \imgfeatureset \rightarrow \textfeatureset$, where $\translator$ can be a simple multi-layer perceptron (MLP), using the correspondences ($\sentence_j$, $\image_i$) and the following objectives.

\vspace{-1em}
\paragraph{Robust Alignment.}
If we train the alignment using a simple $L_2 = \sum_j \| \translator(\imgfeature_i) - \textfeature_j \|_2^2$ loss 
the optimal mapping $\translator^*$ would be the conditional average $\translator^*(\imgfeature_i) \!=\! \sum_j p(\textfeature_j \!\mid\! \image_i)\;\textfeature_j$ which might not be an optimal or verbally rich sentence embedding as it could land between modes of the distribution.
Thus, we propose to learn the feature alignment using a robust formulation that encourages the mapping to be close to a real sentence embedding:
\begin{equation}
    L_R(\imgfeature_i) = \min_{\textfeature_j \sim p(\sentence_j | \image_i)} \|\, \translator(\imgfeature_i) - \textfeature_j \,\|_2^2 \,.
    \label{eq:robust_loss}
\end{equation}
Since the set of matches is very large, we approximate the loss by sampling a fixed amount $K$ of $\textfeature_j$ for each image and by computing the minimum in this subset. 
 
\vspace{-1em}
\paragraph{Adversarial Training.}
So far, the robust alignment encourages to learn a translation $\translator$ that adheres to the structure of the conceptual text embedding. However, we need to ensure that the mapping does not discard important concept information from the image feature vector. This is necessary so that the decoder can decode a caption that directly corresponds to the visual concepts in the image. To this end, we employ adversarial training using a conditional discriminator. 
Since adversarial training on discrete sequences is problematic \cite{chen2018adversarial,subramanian2018towards}, we perform it in feature space $\textfeatureset$ similar to~\cite{subramanian2018towards}. The discriminator $\discriminator: \textfeatureset\times\conceptset \rightarrow \real$ is trained with a set of positive/real and a set of negative/fake examples.
In our case a positive example is the concatenation of a translated feature $\translator(\imgfeature_i)$ with the one-hot encoding of the image concepts $\visconceptset_i$. 
A negative example analogously is the concatenation of the sampled pair's text embedding $\textfeature_j$ and the image concepts $\visconceptset_i$.
Thus, the discriminator learns the correlation of image concepts and text embeddings, which in turn encourages the mapping $\translator$ to encode image concepts correctly. Otherwise the discriminator can easily identify a real sentence feature from a translated image feature. 

In practice, we use a WGAN-GP formulation \cite{gulrajani2017improved} to train the discriminator $\discriminator$ to maximize its output for fake examples and minimize it for real.
When training $\translator$ we thus maximize the discriminator for the translation.
\begin{equation}
    L_{\operatorname{adv}} = - \discriminator(\translator(\imgfeature_i), \, \visconceptset_i)
\end{equation}

\paragraph{Total loss.} Our final model is trained with all three aforementioned objectives:
\begin{equation}
    L_{total} = \lambda_{CE}\,L_{\operatorname{CE}} + \lambda_R\,L_R + \lambda_{\operatorname{adv}}\,L_{\operatorname{adv}},
\end{equation}

where the weight factors $\lambda_{CE}, \lambda_R, \lambda_{\operatorname{adv}} \in \real$ balance the contributions of the three losses.

\section{Experiments and Results}
The evaluation is structured as follows. 
First, we present ablation experiments in an \textit{unpaired} setting on Microsoft COCO~\cite{lin2014microsoft} to evaluate the effect of each component of our method. 
Second, we report the results in the \textit{unsupervised} setting with independent image and language sources. 
We experiment with Flickr30k Images~\cite{plummer2015flickr30k} paired with COCO captions and COCO images paired with Google's Conceptual Captions dataset (GCC)~\cite{sharma2018conceptual}.
Finally, we show qualitative results for image descriptions with varying text sources.

\vspace{-1em}
\paragraph{Implementation details.}
We tokenize and process all natural language datasets, replacing the least frequently used words with \texttt{unk} tokens. The next step is to extract visual word synsets. We use the Visual Genome~\cite{krishna2017visual} object synsets as reference and look up nouns (or noun phrases) extracted by parsing each sentence with the Stanford CoreNLP toolkit~\cite{manning2014stanford}. This results in 1415 synsets for COCO and 3030 synsets for GCC which describe visual entities. During the semantic-aware training of the language model with Equation~\ref{eq:triplet_loss}, positive and negative pairs of captions are defined using this synset vocabulary.

The encoder and decoder of the language model are implemented using Gated Recurrent Units (GRUs)~\cite{cho2014learning} with 200 hidden units. 
The last hidden state of the encoder is projected through a linear layer into 256-d text features $\textfeature$. 
The decoder is followed by a linear layer that maps its output into a fixed-size vocabulary vector. 
We use 200-d GloVe embeddings~\cite{pennington2014glove} as inputs to the language model.

\begin{table*}[!ht]
    \centering
    \begin{tabular}{l @{\hskip2pt}| c@{\hskip5pt} c@{\hskip5pt} c@{\hskip5pt} c@{\hskip5pt} | c c c c@{\hskip8pt} c@{\hskip7pt} c c c c}
        \toprule
        \multicolumn{5}{c}{Component Evaluation} & \multicolumn{9}{c}{Metrics} \\
        \cmidrule{1-5}
        \cmidrule{8-12}
        Abbreviation & $L_{\operatorname{CE}}$ & $L_{2}$ & $L_{R}$ & $L_{\operatorname{adv}}$ & \small{B-1} & \small{B-2} & \small{B-3} & \small{B-4} & \small{METEOR} & \small{ROUGE} & \small{CIDER} & \small{SPICE} & \small{WMD}\\
        \midrule
        \rowcolor{Gray}
         Supervised baseline & & & & & 67.4 & 50.0 & 35.4 & 24.8 & 22.6 & 50.1 & 80.2 & 15.9 & 17.9 \\
         \midrule
         Oracle & & & & & 49.1 & 31.2 & 21.2 & 16.0 & 18.7 & 38.7 & 50.4 & 12.2 & 14.5 \\
         \midrule
         Alignment only &  & \checkmark & & & 47.0 & 25.4 & 11.5 & 5.2 & 15.5 & 35.9 & 29.4 & 8.7 & 9.1\\
         MLE only & \checkmark & & & & 59.9 & 40.2 & 26.0 & 17.1 & 19.1 & 43.7 & 57.9 & 11.6 & 13.0\\
         Joint, baseline & \checkmark & \checkmark & & & 59.7 & 40.2 & 25.8 & 16.6 & 18.3 & 43.1 & 53.8 & 10.8 & 12.6\\
         Joint, robust & \checkmark &  & \checkmark & & 61.5 & 42.3 & 28.0 & 18.8 & 19.7 & 44.9 & 62.4 & 12.5 & 14.3\\
         Joint, robust ($\lambda_t\!=\!0$) & \checkmark &  & \checkmark & & 60.7 & 41.1 & 26.7 & 17.6 & 18.3 & 43.8 & 55.6 & 11.0 & 13.0\\
         Joint, adversarial & \checkmark &  & \checkmark & \checkmark & \textbf{61.7} & \textbf{42.8} & \textbf{28.6} & \textbf{19.3} & \textbf{20.1} & \textbf{45.4} & \textbf{63.6} & \textbf{12.8} & \textbf{14.4}\\
         \bottomrule
    \end{tabular}
    \vspace{0.2em}
    \caption{Ablation Experiments on COCO test set~\cite{karpathy2015deep}. Image and language data are unpaired; COCO ground truth object categories are used for the initial alignment. Every component of our domain alignment model improves the performance on the captioning task. }
    \label{tab:ablation}
\end{table*}

Similar to sentence pairs, we build weak image-sentence assignments based on (visual) synsets to train the image captioner. 
For richness in visual concepts, we use the OpenImages-v4 dataset~\cite{OpenImages2,OpenImages}, which consists of 1.74 million images and 600 annotated object categories.
Visual concepts are extracted using a Faster R-CNN detector~\cite{huang2017speed} trained on OpenImages, which has been made publicly available\footnote{\url{https://github.com/tensorflow/models/tree/master/research/object_detection}}. 
Please note that we only make use of class labels and do not rely on image regions (bounding boxes) in order to keep the amount of supervision minimal. Thus, any multi-label classifier could be used instead. 

The baseline for our image captioner is based on \cite{vinyals2015show} and uses image features extracted by ResNet-101~\cite{he2016deep} pretrained on ImageNet, without finetuning.
The translator $\translator$ is implemented with a single-layer MLP of size 512 to map $\imgfeature \in \real^{2048}$ into $\textfeature \in \real^{256}$. 

\vspace{-1em}
\paragraph{Training details.}
We train the language model until convergence with a batch size of 64. 
The initial learning rates of the encoder and decoder are set to $10^{-4}$ and $10^{-3}$ respectively and $\lambda_{t} = 0.1$.
When training the the alignment model, we further finetune the decoder so that it adapts to the joint embedding space. We optimize using Adam~\cite{kingma2014adam} with a learning rate of $10^{-3}$ and $\lambda_{CE}\!=\!\lambda_R\!=\!1$, $\lambda_{\operatorname{adv}}\!=\!0.1$. 
 
\vspace{-1em}
\paragraph{Evaluation metrics.}
We evaluate our method with the official COCO evaluation code and report performance under the commonly used metrics, BLEU 1-4~\cite{cho2014learning}, ROUGE~\cite{lin2004rouge}, METEOR~\cite{denkowski2014meteor}, CIDEr~\cite{vedantam2015cider}, SPICE~\cite{anderson2016spice} and WMD~\cite{kusner2015word}.

\subsection{Unpaired Captioning}
The unpaired setting on COCO allows us to evaluate the effectiveness of the proposed method and to compare to previous work~\cite{feng2018unsupervised} using the same controlled setup.
This is a simplification of the problem since the images and their descriptions come from the same distribution; however, we do not use the ground truth correspondences and treat images and text unpaired. 
We use the same data splits %
as in previous methods following \cite{karpathy2015deep}, resulting in 113,287 training, 5,000 validation and 5,000 test images. Each image is originally annotated with 5 descriptions, resulting in over 560k training captions. 
After generating our initial image-caption assignments based on visual synsets, there are approximately 150k unique captions remaining in the graph $\graph$. 

\vspace{-1em}
\paragraph{Ablation study.}
We evaluate the proposed components through ablation experiments (Table~\ref{tab:ablation}).
In these experiments, we use the 80 available COCO object categories as visual concepts. 
We compare the following models. 

\emph{Oracle:}
We first evaluate the weak assignments using an oracle that selects the highest probability candidate among the ground truth captions assigned to an image. 
This candidate has the highest overlap of visual concepts with the image.    
Since there can be multiple captions with equally high probability, we randomly sample and report the best out of 100 runs. 
This baseline scores generally low as the initial assignments are very noisy.

\emph{Alignment only:}
The alignment is performed by training only the mapping $\translator$ of image features into the sentence manifold. 
We keep the decoder frozen, using the weights from the pretrained language model. 
The model shows understanding of the major visual concepts in the scene, meaning that relevant classes appear in the output sentence. 
However, the sentences are grammatically incoherent because the decoder cannot adapt to the latent space difference between the projected image features and the real sentence embeddings 
it was trained with.
Thus, for subsequent experiments we also jointly finetune the decoder. 

\emph{MLE only:} 
The full model is trained using the weak pairs of image-captions and teacher forcing, in the standard supervised manner, but without any constraints to encourage a shared domain. The model is prone to the bias often seen in MLE models such as repeating sub-phrases. 

\emph{Joint, baseline:}
In addition to MLE training, domain alignment is performed by minimizing the $L_2$-distance between $\translator(\imgfeature)$ and $\textfeature$. 
This na\"ive alignment of the two domains does not improve over the MLE-only baseline. 

\emph{Joint, robust:}
Instead of L2, the model is trained with the proposed robust alignment loss \eqref{eq:robust_loss} which gives a significant boost in performance. We randomly sample $K=10$ sentences as candidate pairs for each training image.  

\emph{Joint, robust ($\lambda_t\!=\!0$):}
To evaluate the importance of the embedding space, we also train the above model against sentence embeddings that come from a language model trained only with $L_{\operatorname{LM}}\!\!\!:=\!\!\!L_{\operatorname{CE}}$, \ie without the triplet loss.    
It performs worse, suggesting that the semantic structure of the language model is indeed beneficial for captioning. 

\emph{Joint, adversarial:}
The full model additionally includes adversarial training conditioned on visual concepts as categorical inputs. We observe that our unpaired model reaches performance close to its fully supervised counterpart~\cite{vinyals2015show} and is comparable to early work on image captioning. 

The consistent improvement shows that our model is able to learn concepts beyond the initial weak assignments.

\begin{figure*}[t]
	\centering
	\includegraphics[clip, trim=0cm 3cm 0cm 0cm, width=1.0\textwidth]{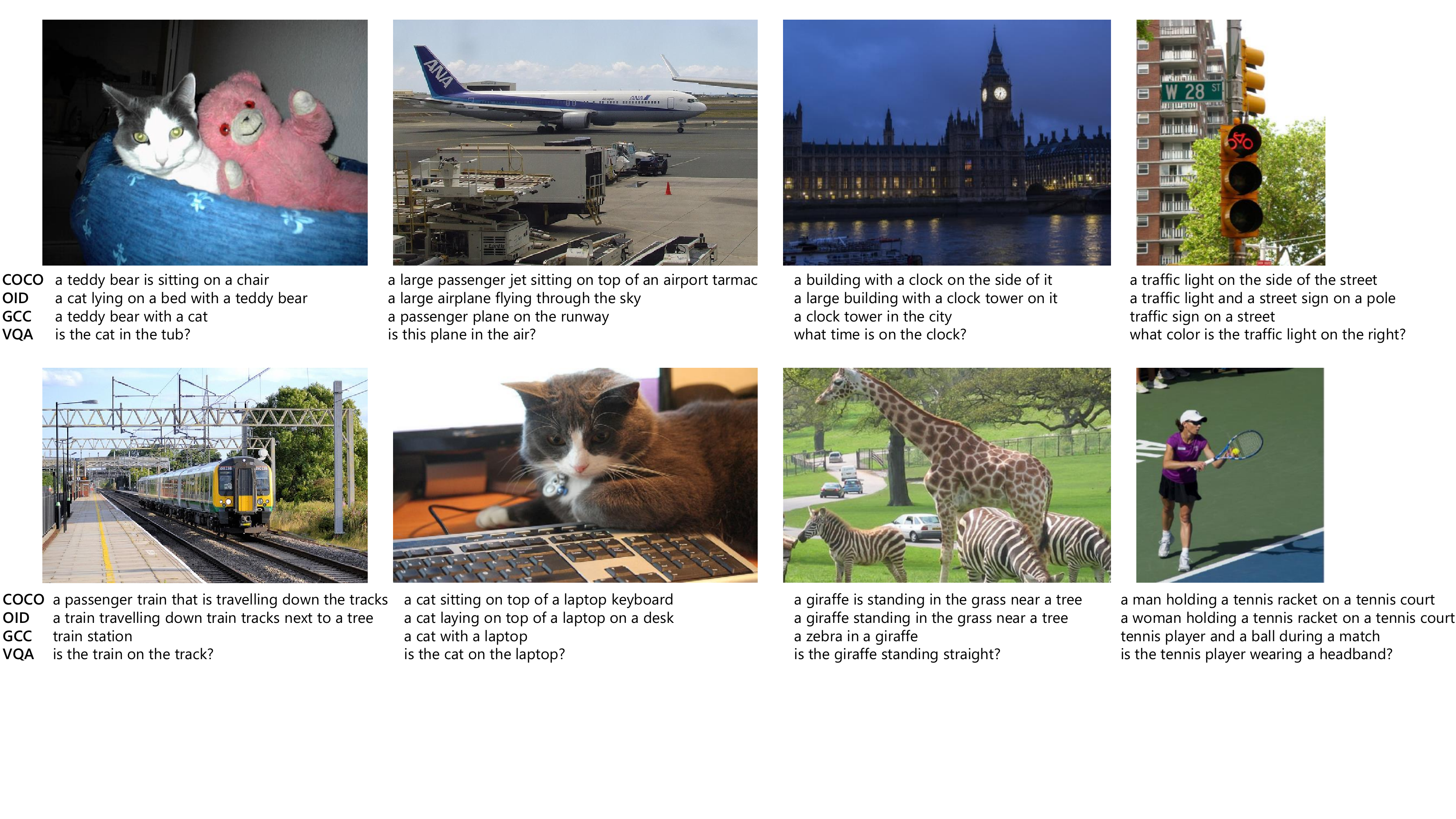}
	\caption{\textbf{Qualitative Results.} We show caption predictions on images from the COCO dataset. COCO and OID are results from our \textit{unpaired} model trained with weak pairs coming from a detector trained on the respective dataset. GCC and VQA refer to the \textit{unsupervised} model trained on COCO images using the Conceptual Captions and VQA-v2 datasets respectively. 
	}
	\label{fig:results}
\end{figure*}

\vspace{-1em}
\paragraph{Comparison to the State of the Art.}
The field of image captioning without image-caption pairs has only been explored very recently. In Table~\ref{tab:comparison_sota}, we compare our approach to previous methods. We follow the same \textit{unpaired} setup on COCO as in \cite{feng2018unsupervised}. 
We use the object detector trained on OpenImages (OID) to predict visual concepts for both creating the image-caption assignments and conditioning the discriminator during adversarial training. The reported results correspond to the predictions from our full model trained with $K=10$ samples and evaluated using a beam size of $3$. Our method sets a new state of the art on this problem. 

\vspace{-1em}
\paragraph{Qualitative Evaluation.}
We show qualitative results of our full model in Figure~\ref{fig:results}, comparing captions predicted in the unpaired setting with two variants trained with different visual concept extractors (COCO and OID). 
We find that both the COCO model and the OID model capture the image contents well, whereas the OID model clearly benefits from the richer object detections. 
For example, in the last image the COCO model produces a description about a \textit{man} ---potentially due to bias. 
This is because only \textit{person} is a category in COCO, but not \textit{man} or \textit{woman}, and therefore there can be no gender distinction in the captions that are weakly assigned to each image. 
The model trained with OID concepts has the capacity to resolve such ambiguities and correctly identifies \textit{woman} in the last image. 
We note that the object detector is only used during training (for the weak assignments and the discriminator), but not during inference.
The captioner learns to extrapolate from the labeled categories of the image domain; \eg the generated words \{tracks, airport, tower, passenger, grass\} are unlabeled concepts that the model inferred due to co-occurrence with labeled concepts such as train, airplane, clock, etc. 

\subsection{Unsupervised Captioning}
When training the image captioner in an unsupervised manner, the language model is pre-trained using an external text source and all other settings remain identical. 
We perform two cross-domain experiments: COCO images with GCC sentences and Flickr30k images with COCO captions. Quantitative results can be seen in Table~\ref{tab:datasets} for the model variants with and without adversarial training. Adversarial training consistently improves our model.
Naturally, we do not expect to match the performance of the unpaired setting since a different language domain implies vocabulary, context and style that differs from the \textit{ground truth} captions in 
COCO. 

Qualitatively, we show the predicted captions of the model trained on COCO images and GCC captions in Figure~\ref{fig:results} (denoted as GCC).
When using GCC as the language domain, we find that the initial image-caption assignments are even more noisy, which leads the model to produce short and simple descriptions.
However, we also see that this model has learned some interesting concepts, not present in the unpaired setting, such as the difference between a plane being on the ground or in the air.
 
To produce descriptions with different styles that extend beyond captioning datasets, the choice of the language domain is not trivial, as it should be rich in visual descriptions.
We thus experiment with VQA-v2~\cite{antol2015vqa} as the language domain, using the questions provided by the dataset as the sentence source. Instead of captioning, the model learns to ask questions about the image content (Figure~\ref{fig:results}, VQA).  

\subsection{Joint Embedding Visualization}
\begin{figure}[th]
    \centering
    \includegraphics[clip, trim=0cm 10cm 6.5cm 0.4cm, width=1.00\columnwidth]{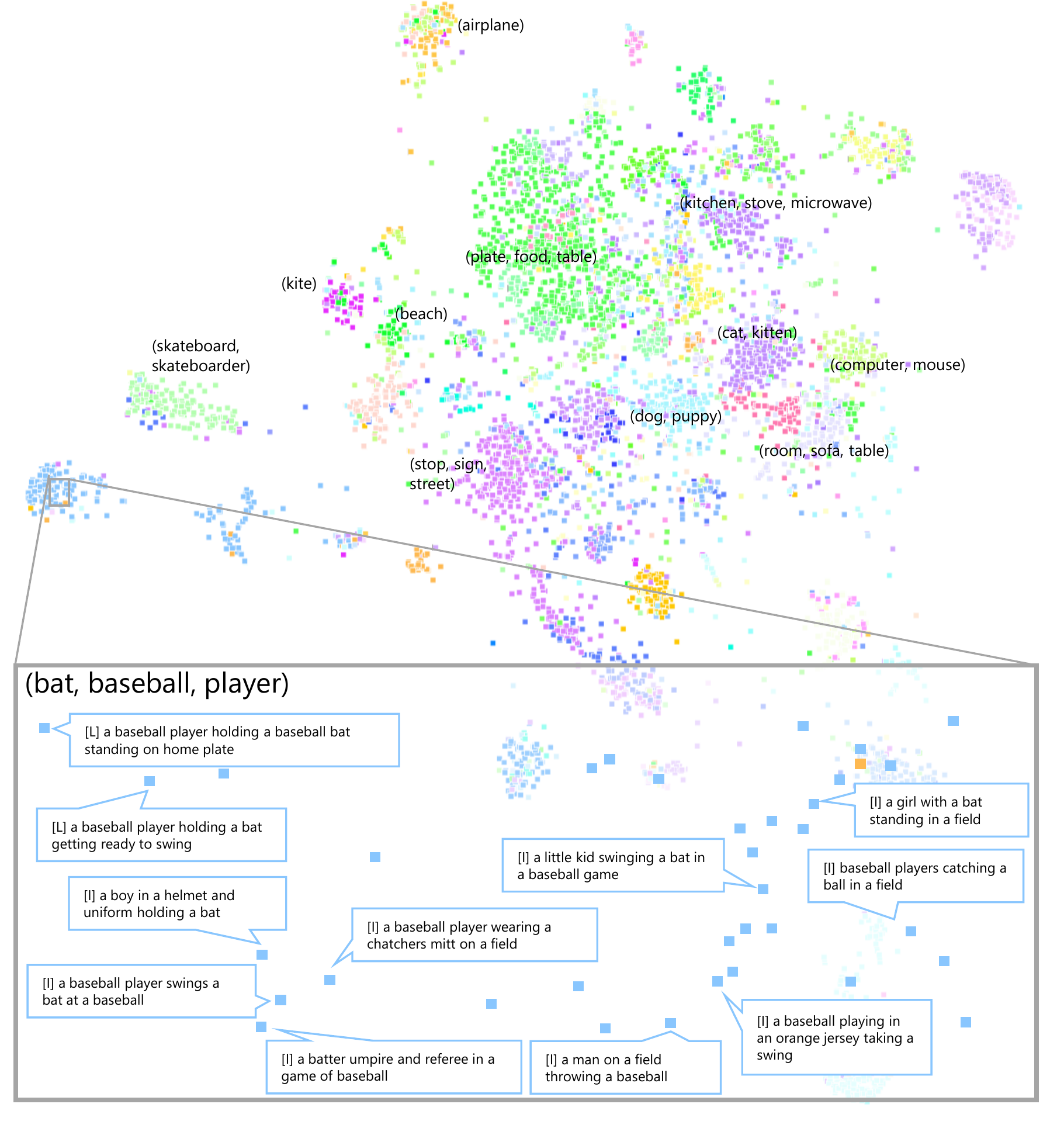}
    \caption{\textbf{$t$-SNE Embedding.} We show a projection of the learned joint embedding of our model 
    and zoom into a cluster to visualize that sentences from the text corpus (denoted by [L]) lie in visual-semantic groups together with image embeddings [I]. Colors are generated by groups of visual concepts. A large-scale version of this figure can be found in the supplementary material.}
    \label{fig:tsne}
\end{figure}

Finally, to verify that our training creates a meaningful \textit{joint} latent space, we visualize the $t$-SNE embedding \cite{maaten2008visualizing} of both the sentences (marked with [L]) and image-projected features ([I]) in Figure \ref{fig:tsne}. The overall embedding is structured by visual categories due to the constraints we impose on the model during training. Within clusters, image and text features are well mixed. This means that the model has learned a joint embedding where it is not possible to separate text form images. 

\begin{table}[t]
	\centering
	\begin{tabular}{l c c c c c}
		\toprule
		\multirow{2}{*}{Method} & \multicolumn{5}{c}{Metrics} \\ 
		\cmidrule{2-6}
		& B-4 & M & R & C & S \\
		\midrule
		Gu~\etal~\cite{gu2018unpaired} & 5.4 & 13.2 & - & 17.7 & - \\
		Feng~\etal~\cite{feng2018unsupervised} & 18.6 & 17.9 & 43.1 & 54.9 & 11.1 \\
		\midrule
		Ours & \textbf{19.3} & \textbf{20.2} & \textbf{45.0} & \textbf{61.8} & \textbf{12.9}\\
		\bottomrule
	\end{tabular}
	\vspace{1.0em}
	\caption{Comparison with the state of the art on COCO test set~\cite{karpathy2015deep} under the \textit{unpaired} setting of \cite{feng2018unsupervised}. OpenImages~\cite{OpenImages2} categories are used for concept extraction.}
	\label{tab:comparison_sota}
\end{table}

\begin{table}[t]
	\centering
	\begin{tabular}{l c c c c c}
		\toprule
		\multirow{2}{*}{Method} & \multicolumn{5}{c}{Metrics} \\ 
		\cmidrule{2-6}
		& B-4 & M & R & C & W \\
		\midrule
		\multicolumn{6}{l}{\textbf{Flickr Images $\leftrightarrow$ COCO Captions}} \\ \addlinespace[0.25em]
		Ours (w/o adv) & 5.9 & 10.9 & 31.1 & 8.2 & 7.0\\
		Ours & 7.9 & 13.0 & 32.8 & 9.9 & 7.5 \\   
		\midrule
		\multicolumn{6}{l}{\textbf{COCO Images $\leftrightarrow$ Conceptual Captions}}\\ \addlinespace[0.25em]
		Ours (w/o adv) & 5.5 & 11.1 & 30.1 & 20.8 & 6.7\\
		Ours & 6.5 & 12.9 & 35.1 & 22.7 & 7.4 \\   
		\bottomrule
	\end{tabular}
	\vspace{1.0em}
	\caption{Evaluation under the unsupervised setting using image and captions from independent sources.}
	\label{tab:datasets}
\end{table}

\section{Limitations and Discussion}
\label{sec:discussion}

Although our approach sets the state of the art in unsupervised image captioning, there are still several limitations. 
As mentioned before, to generate the initial assignments, the language source needs to contain sufficient visual concepts overlapping with the image domain. We believe it is possible to alleviate this problem by learning from a combination of text sources with varying contents and styles.

Another limitation is the capability of the model to extend to novel compositions and atypical scene descriptions. We observe two factors that decide the model's behavior in this respect. First, the capabilities of the base captioner itself, \ie unsupervised training will not solve limitations that are present even for the supervised model~\cite{vinyals2015show}. In our experiments, the output often collapses into caption modes that are generic enough to describe a set of images; this results in approximately 20\% of the generated captions actually being unique and 16\% novel captions, not found in the training set. This is on par with the findings of \cite{vinyals2015show}.  

The second factor is the amount of \textit{discoverable} visual concepts. 
For example, it is not possible to discover the difference between a whole pizza and a slice of pizza, when only the concept \textit{pizza} is known, unless \textit{slice} also appears in other context. 
Naturally, learning from more concepts holds the potential for more diversity. 
One could enrich the search space of weak assignments by including predicates in the set of known visual concepts, thus relying on relationship detection.
This could greatly help in resolving ambiguities such as \textit{a person riding a bike} or \textit{carrying a bike}, however it goes against the idea of weak or no supervision.

\section{Conclusion}
We have presented a novel method to align images and text in a shared latent repesentation that is structured through visual concepts. Our method is minimally supervised in the sense that it requires a standard, pre-trained image recognition model to obtain initial noisy correspondences between the image and the text domain. 
Our robust training scheme and the adversarial learning of the translation from image features to text allows the model to successfully learn the captioning task. In our experiments we show different combinations of image and text sources and improve the state of the art in the unpaired COCO setting.

For the future we are interested in investigating several directions. One could improve the decoder architecture with typical components, such as attention, 
or follow a template approach to encourage novel compositions of objects. 
Overall, unsupervised image captioning is an upcoming research direction that is gaining traction in the community. 

\vspace{-0.5em}
\paragraph{Acknowledgements.}
Christian Rupprecht is supported by ERC Stg Grant IDIU-638009.

{\small
	\bibliographystyle{ieee_fullname}
	\bibliography{references}

\begin{thebibliography}{10}\itemsep=-1pt

\bibitem{anderson2016spice}
Peter Anderson, Basura Fernando, Mark Johnson, and Stephen Gould.
\newblock Spice: Semantic propositional image caption evaluation.
\newblock In {\em European Conference on Computer Vision}, pages 382--398.
  Springer, 2016.

\bibitem{Anderson2017GuidedOV}
Peter Anderson, Basura Fernando, Mark Johnson, and Stephen Gould.
\newblock Guided open vocabulary image captioning with constrained beam search.
\newblock In {\em EMNLP}, 2017.

\bibitem{anderson2018partially}
Peter Anderson, Stephen Gould, and Mark Johnson.
\newblock Partially-supervised image captioning.
\newblock In {\em Advances in Neural Information Processing Systems}, pages
  1879--1890, 2018.

\bibitem{anderson2018bottom}
Peter Anderson, Xiaodong He, Chris Buehler, Damien Teney, Mark Johnson, Stephen
  Gould, and Lei Zhang.
\newblock Bottom-up and top-down attention for image captioning and visual
  question answering.
\newblock In {\em Proceedings of the IEEE Conference on Computer Vision and
  Pattern Recognition}, pages 6077--6086, 2018.

\bibitem{anne2018women}
Lisa Anne~Hendricks, Kaylee Burns, Kate Saenko, Trevor Darrell, and Anna
  Rohrbach.
\newblock Women also snowboard: Overcoming bias in captioning models.
\newblock In {\em Proceedings of the European Conference on Computer Vision
  (ECCV)}, pages 771--787, 2018.

\bibitem{anne2016deep}
Lisa Anne~Hendricks, Subhashini Venugopalan, Marcus Rohrbach, Raymond Mooney,
  Kate Saenko, and Trevor Darrell.
\newblock Deep compositional captioning: Describing novel object categories
  without paired training data.
\newblock In {\em Proceedings of the IEEE conference on computer vision and
  pattern recognition}, pages 1--10, 2016.

\bibitem{antol2015vqa}
Stanislaw Antol, Aishwarya Agrawal, Jiasen Lu, Margaret Mitchell, Dhruv Batra,
  C.~Lawrence Zitnick, and Devi Parikh.
\newblock {VQA}: {V}isual {Q}uestion {A}nswering.
\newblock In {\em International Conference on Computer Vision (ICCV)}, 2015.

\bibitem{chen2018adversarial}
Liqun Chen, Shuyang Dai, Chenyang Tao, Haichao Zhang, Zhe Gan, Dinghan Shen,
  Yizhe Zhang, Guoyin Wang, Ruiyi Zhang, and Lawrence Carin.
\newblock Adversarial text generation via feature-mover's distance.
\newblock In {\em Advances in Neural Information Processing Systems}, pages
  4671--4682, 2018.

\bibitem{chen2017show}
Tseng-Hung Chen, Yuan-Hong Liao, Ching-Yao Chuang, Wan-Ting Hsu, Jianlong Fu,
  and Min Sun.
\newblock Show, adapt and tell: Adversarial training of cross-domain image
  captioner.
\newblock In {\em Proceedings of the IEEE International Conference on Computer
  Vision}, pages 521--530, 2017.

\bibitem{cho2014learning}
Kyunghyun Cho, Bart Van~Merri{\"e}nboer, Caglar Gulcehre, Dzmitry Bahdanau,
  Fethi Bougares, Holger Schwenk, and Yoshua Bengio.
\newblock Learning phrase representations using {RNN} encoder-decoder for
  statistical machine translation.
\newblock {\em arXiv preprint arXiv:1406.1078}, 2014.

\bibitem{dai2017towards}
Bo Dai, Sanja Fidler, Raquel Urtasun, and Dahua Lin.
\newblock Towards diverse and natural image descriptions via a conditional
  {GAN}.
\newblock In {\em Proceedings of the IEEE International Conference on Computer
  Vision}, pages 2970--2979, 2017.

\bibitem{das2017visual}
Abhishek Das, Satwik Kottur, Khushi Gupta, Avi Singh, Deshraj Yadav,
  Jos{\'e}~MF Moura, Devi Parikh, and Dhruv Batra.
\newblock Visual dialog.
\newblock In {\em Proceedings of the IEEE Conference on Computer Vision and
  Pattern Recognition}, pages 326--335, 2017.

\bibitem{denkowski2014meteor}
Michael Denkowski and Alon Lavie.
\newblock Meteor universal: Language specific translation evaluation for any
  target language.
\newblock In {\em Proceedings of the ninth workshop on statistical machine
  translation}, pages 376--380, 2014.

\bibitem{donahue2015long}
Jeffrey Donahue, Lisa Anne~Hendricks, Sergio Guadarrama, Marcus Rohrbach,
  Subhashini Venugopalan, Kate Saenko, and Trevor Darrell.
\newblock Long-term recurrent convolutional networks for visual recognition and
  description.
\newblock In {\em Proceedings of the IEEE conference on computer vision and
  pattern recognition}, pages 2625--2634, 2015.

\bibitem{donahue2016adversarial}
Jeff Donahue, Philipp Kr{\"a}henb{\"u}hl, and Trevor Darrell.
\newblock Adversarial feature learning.
\newblock {\em arXiv preprint arXiv:1605.09782}, 2016.

\bibitem{faghri2017vse++}
Fartash Faghri, David~J Fleet, Jamie~Ryan Kiros, and Sanja Fidler.
\newblock {VSE++}: Improving visual-semantic embeddings with hard negatives.
\newblock {\em arXiv preprint arXiv:1707.05612}, 2017.

\bibitem{fang2015captions}
Hao Fang, Saurabh Gupta, Forrest Iandola, Rupesh~K Srivastava, Li Deng, Piotr
  Doll{\'a}r, Jianfeng Gao, Xiaodong He, Margaret Mitchell, John~C Platt,
  et~al.
\newblock From captions to visual concepts and back.
\newblock In {\em Proceedings of the IEEE conference on computer vision and
  pattern recognition}, pages 1473--1482, 2015.

\bibitem{feng2018unsupervised}
Yang Feng, Lin Ma, Wei Liu, and Jiebo Luo.
\newblock Unsupervised image captioning.
\newblock {\em arXiv preprint arXiv:1811.10787}, 2018.

\bibitem{gan2017stylenet}
Chuang Gan, Zhe Gan, Xiaodong He, Jianfeng Gao, and Li Deng.
\newblock {StyleNet}: Generating attractive visual captions with styles.
\newblock In {\em Proceedings of the IEEE Conference on Computer Vision and
  Pattern Recognition}, pages 3137--3146, 2017.

\bibitem{gu2018unpaired}
Jiuxiang Gu, Shafiq Joty, Jianfei Cai, and Gang Wang.
\newblock Unpaired image captioning by language pivoting.
\newblock In {\em Proceedings of the European Conference on Computer Vision
  (ECCV)}, pages 503--519, 2018.

\bibitem{gulrajani2017improved}
Ishaan Gulrajani, Faruk Ahmed, Martin Arjovsky, Vincent Dumoulin, and Aaron~C
  Courville.
\newblock Improved training of wasserstein gans.
\newblock In {\em Advances in Neural Information Processing Systems}, pages
  5767--5777, 2017.

\bibitem{Guo_2019_CVPR}
Longteng Guo, Jing Liu, Peng Yao, Jiangwei Li, and Hanqing Lu.
\newblock {MSCap}: Multi-style image captioning with unpaired stylized text.
\newblock In {\em The IEEE Conference on Computer Vision and Pattern
  Recognition (CVPR)}, June 2019.

\bibitem{gurari2018vizwiz}
Danna Gurari, Qing Li, Abigale~J Stangl, Anhong Guo, Chi Lin, Kristen Grauman,
  Jiebo Luo, and Jeffrey~P Bigham.
\newblock {VizWiz} grand challenge: Answering visual questions from blind
  people.
\newblock In {\em Proceedings of the IEEE Conference on Computer Vision and
  Pattern Recognition}, pages 3608--3617, 2018.

\bibitem{he2016deep}
Kaiming He, Xiangyu Zhang, Shaoqing Ren, and Jian Sun.
\newblock Deep residual learning for image recognition.
\newblock In {\em Proceedings of the IEEE conference on computer vision and
  pattern recognition}, pages 770--778, 2016.

\bibitem{huang2017speed}
Jonathan Huang, Vivek Rathod, Chen Sun, Menglong Zhu, Anoop Korattikara,
  Alireza Fathi, Ian Fischer, Zbigniew Wojna, Yang Song, Sergio Guadarrama,
  et~al.
\newblock Speed/accuracy trade-offs for modern convolutional object detectors.
\newblock In {\em Proceedings of the IEEE conference on computer vision and
  pattern recognition}, pages 7310--7311, 2017.

\bibitem{johnson2016densecap}
Justin Johnson, Andrej Karpathy, and Li Fei-Fei.
\newblock {DenseCap}: Fully convolutional localization networks for dense
  captioning.
\newblock In {\em Proceedings of the IEEE Conference on Computer Vision and
  Pattern Recognition}, pages 4565--4574, 2016.

\bibitem{karpathy2015deep}
Andrej Karpathy and Li Fei-Fei.
\newblock Deep visual-semantic alignments for generating image descriptions.
\newblock In {\em Proceedings of the IEEE conference on computer vision and
  pattern recognition}, pages 3128--3137, 2015.

\bibitem{kingma2014adam}
Diederik~P Kingma and Jimmy Ba.
\newblock Adam: A method for stochastic optimization.
\newblock {\em arXiv preprint arXiv:1412.6980}, 2014.

\bibitem{kiros2014unifying}
Ryan Kiros, Ruslan Salakhutdinov, and Richard~S Zemel.
\newblock Unifying visual-semantic embeddings with multimodal neural language
  models.
\newblock {\em arXiv preprint arXiv:1411.2539}, 2014.

\bibitem{OpenImages2}
Ivan Krasin, Tom Duerig, Neil Alldrin, Vittorio Ferrari, Sami Abu-El-Haija,
  Alina Kuznetsova, Hassan Rom, Jasper Uijlings, Stefan Popov, Shahab Kamali,
  Matteo Malloci, Jordi Pont-Tuset, Andreas Veit, Serge Belongie, Victor Gomes,
  Abhinav Gupta, Chen Sun, Gal Chechik, David Cai, Zheyun Feng, Dhyanesh
  Narayanan, and Kevin Murphy.
\newblock {OpenImages}: A public dataset for large-scale multi-label and
  multi-class image classification.
\newblock {\em Dataset available from
  https://storage.googleapis.com/openimages/web/index.html}, 2017.

\bibitem{krishna2017visual}
Ranjay Krishna, Yuke Zhu, Oliver Groth, Justin Johnson, Kenji Hata, Joshua
  Kravitz, Stephanie Chen, Yannis Kalantidis, Li-Jia Li, David~A Shamma, et~al.
\newblock Visual genome: Connecting language and vision using crowdsourced
  dense image annotations.
\newblock {\em International Journal of Computer Vision}, 123(1):32--73, 2017.

\bibitem{kusner2015word}
Matt Kusner, Yu Sun, Nicholas Kolkin, and Kilian Weinberger.
\newblock From word embeddings to document distances.
\newblock In {\em International Conference on Machine Learning}, pages
  957--966, 2015.

\bibitem{OpenImages}
Alina Kuznetsova, Hassan Rom, Neil Alldrin, Jasper Uijlings, Ivan Krasin, Jordi
  Pont-Tuset, Shahab Kamali, Stefan Popov, Matteo Malloci, Tom Duerig, and
  Vittorio Ferrari.
\newblock The open images dataset v4: Unified image classification, object
  detection, and visual relationship detection at scale.
\newblock {\em arXiv:1811.00982}, 2018.

\bibitem{lample2017unsupervised}
Guillaume Lample, Alexis Conneau, Ludovic Denoyer, and Marc'Aurelio Ranzato.
\newblock Unsupervised machine translation using monolingual corpora only.
\newblock In {\em International Conference on Learning Representations (ICLR)},
  2018.

\bibitem{lample2018phrase}
Guillaume Lample, Myle Ott, Alexis Conneau, Ludovic Denoyer, and Marc'Aurelio
  Ranzato.
\newblock Phrase-based \& neural unsupervised machine translation.
\newblock In {\em Proceedings of the 2018 Conference on Empirical Methods in
  Natural Language Processing (EMNLP)}, 2018.

\bibitem{li2018generating}
Dianqi Li, Xiaodong He, Qiuyuan Huang, Ming-Ting Sun, and Lei Zhang.
\newblock Generating diverse and accurate visual captions by comparative
  adversarial learning.
\newblock {\em arXiv preprint arXiv:1804.00861}, 2018.

\bibitem{lin2004rouge}
Chin-Yew Lin.
\newblock Rouge: A package for automatic evaluation of summaries.
\newblock {\em Text Summarization Branches Out}, 2004.

\bibitem{lin2014microsoft}
Tsung-Yi Lin, Michael Maire, Serge Belongie, James Hays, Pietro Perona, Deva
  Ramanan, Piotr Doll{\'a}r, and C~Lawrence Zitnick.
\newblock Microsoft {COCO}: Common objects in context.
\newblock In {\em European conference on computer vision}, pages 740--755.
  Springer, 2014.

\bibitem{ling2017teaching}
Huan Ling and Sanja Fidler.
\newblock Teaching machines to describe images via natural language feedback.
\newblock In {\em Proceedings of the 31st International Conference on Neural
  Information Processing Systems}, pages 5075--5085. Curran Associates Inc.,
  2017.

\bibitem{liu2017improved}
Siqi Liu, Zhenhai Zhu, Ning Ye, Sergio Guadarrama, and Kevin Murphy.
\newblock Improved image captioning via policy gradient optimization of spider.
\newblock In {\em Proceedings of the IEEE international conference on computer
  vision}, pages 873--881, 2017.

\bibitem{lu2017knowing}
Jiasen Lu, Caiming Xiong, Devi Parikh, and Richard Socher.
\newblock Knowing when to look: Adaptive attention via a visual sentinel for
  image captioning.
\newblock In {\em Proceedings of the IEEE conference on computer vision and
  pattern recognition}, pages 375--383, 2017.

\bibitem{lu2018neural}
Jiasen Lu, Jianwei Yang, Dhruv Batra, and Devi Parikh.
\newblock Neural baby talk.
\newblock In {\em Proceedings of the IEEE Conference on Computer Vision and
  Pattern Recognition}, pages 7219--7228, 2018.

\bibitem{maaten2008visualizing}
Laurens van~der Maaten and Geoffrey Hinton.
\newblock Visualizing data using {t-SNE}.
\newblock {\em Journal of machine learning research}, 9(Nov):2579--2605, 2008.

\bibitem{manning2014stanford}
Christopher Manning, Mihai Surdeanu, John Bauer, Jenny Finkel, Steven Bethard,
  and David McClosky.
\newblock The stanford {CoreNLP} natural language processing toolkit.
\newblock In {\em Proceedings of 52nd annual meeting of the association for
  computational linguistics: system demonstrations}, pages 55--60, 2014.

\bibitem{mao2015learning}
Junhua Mao, Xu Wei, Yi Yang, Jiang Wang, Zhiheng Huang, and Alan~L Yuille.
\newblock Learning like a child: Fast novel visual concept learning from
  sentence descriptions of images.
\newblock In {\em Proceedings of the IEEE international conference on computer
  vision}, pages 2533--2541, 2015.

\bibitem{mathews2018semstyle}
Alexander Mathews, Lexing Xie, and Xuming He.
\newblock {SemStyle}: Learning to generate stylised image captions using
  unaligned text.
\newblock In {\em Proceedings of the IEEE Conference on Computer Vision and
  Pattern Recognition}, pages 8591--8600, 2018.

\bibitem{pan2016jointly}
Yingwei Pan, Tao Mei, Ting Yao, Houqiang Li, and Yong Rui.
\newblock Jointly modeling embedding and translation to bridge video and
  language.
\newblock In {\em Proceedings of the IEEE conference on computer vision and
  pattern recognition}, pages 4594--4602, 2016.

\bibitem{pennington2014glove}
Jeffrey Pennington, Richard Socher, and Christopher Manning.
\newblock {GloVe}: Global vectors for word representation.
\newblock In {\em Proceedings of the 2014 conference on empirical methods in
  natural language processing (EMNLP)}, pages 1532--1543, 2014.

\bibitem{plummer2015flickr30k}
Bryan~A Plummer, Liwei Wang, Chris~M Cervantes, Juan~C Caicedo, Julia
  Hockenmaier, and Svetlana Lazebnik.
\newblock Flickr30k entities: Collecting region-to-phrase correspondences for
  richer image-to-sentence models.
\newblock In {\em Proceedings of the IEEE international conference on computer
  vision}, pages 2641--2649, 2015.

\bibitem{rennie2017self}
Steven~J Rennie, Etienne Marcheret, Youssef Mroueh, Jerret Ross, and Vaibhava
  Goel.
\newblock Self-critical sequence training for image captioning.
\newblock In {\em Proceedings of the IEEE Conference on Computer Vision and
  Pattern Recognition}, pages 7008--7024, 2017.

\bibitem{russakovsky2015imagenet}
Olga Russakovsky, Jia Deng, Hao Su, Jonathan Krause, Sanjeev Satheesh, Sean Ma,
  Zhiheng Huang, Andrej Karpathy, Aditya Khosla, Michael Bernstein, et~al.
\newblock {ImageNet} large scale visual recognition challenge.
\newblock {\em International journal of computer vision}, 115(3):211--252,
  2015.

\bibitem{sharma2018conceptual}
Piyush Sharma, Nan Ding, Sebastian Goodman, and Radu Soricut.
\newblock Conceptual captions: A cleaned, hypernymed, image alt-text dataset
  for automatic image captioning.
\newblock In {\em Proceedings of the 56th Annual Meeting of the Association for
  Computational Linguistics (Volume 1: Long Papers)}, volume~1, pages
  2556--2565, 2018.

\bibitem{shetty2017speaking}
Rakshith Shetty, Marcus Rohrbach, Lisa Anne~Hendricks, Mario Fritz, and Bernt
  Schiele.
\newblock Speaking the same language: Matching machine to human captions by
  adversarial training.
\newblock In {\em Proceedings of the IEEE International Conference on Computer
  Vision}, pages 4135--4144, 2017.

\bibitem{shu2018deforming}
Zhixin Shu, Mihir Sahasrabudhe, Riza Alp~Guler, Dimitris Samaras, Nikos
  Paragios, and Iasonas Kokkinos.
\newblock Deforming autoencoders: Unsupervised disentangling of shape and
  appearance.
\newblock In {\em The European Conference on Computer Vision (ECCV)}, September
  2018.

\bibitem{shuster2019engaging}
Kurt Shuster, Samuel Humeau, Hexiang Hu, Antoine Bordes, and Jason Weston.
\newblock Engaging image captioning via personality.
\newblock In {\em Proceedings of the IEEE Conference on Computer Vision and
  Pattern Recognition}, pages 12516--12526, 2019.

\bibitem{subramanian2018towards}
Sandeep Subramanian, Sai~Rajeswar Mudumba, Alessandro Sordoni, Adam Trischler,
  Aaron~C Courville, and Chris Pal.
\newblock Towards text generation with adversarially learned neural outlines.
\newblock In {\em Advances in Neural Information Processing Systems}, pages
  7562--7574, 2018.

\bibitem{sutskever2014sequence}
Ilya Sutskever, Oriol Vinyals, and Quoc~V Le.
\newblock Sequence to sequence learning with neural networks.
\newblock In {\em Advances in neural information processing systems}, pages
  3104--3112, 2014.

\bibitem{vedantam2015cider}
Ramakrishna Vedantam, C Lawrence~Zitnick, and Devi Parikh.
\newblock Cider: Consensus-based image description evaluation.
\newblock In {\em Proceedings of the IEEE conference on computer vision and
  pattern recognition}, pages 4566--4575, 2015.

\bibitem{venugopalan2017captioning}
Subhashini Venugopalan, Lisa Anne~Hendricks, Marcus Rohrbach, Raymond Mooney,
  Trevor Darrell, and Kate Saenko.
\newblock Captioning images with diverse objects.
\newblock In {\em Proceedings of the IEEE Conference on Computer Vision and
  Pattern Recognition}, pages 5753--5761, 2017.

\bibitem{vinyals2015show}
Oriol Vinyals, Alexander Toshev, Samy Bengio, and Dumitru Erhan.
\newblock Show and tell: A neural image caption generator.
\newblock In {\em Proceedings of the IEEE conference on computer vision and
  pattern recognition}, pages 3156--3164, 2015.

\bibitem{wang2017diverse}
Liwei Wang, Alexander Schwing, and Svetlana Lazebnik.
\newblock Diverse and accurate image description using a variational
  auto-encoder with an additive gaussian encoding space.
\newblock In {\em Advances in Neural Information Processing Systems}, pages
  5756--5766, 2017.

\bibitem{wu2017automatic}
Shaomei Wu, Jeffrey Wieland, Omid Farivar, and Julie Schiller.
\newblock Automatic alt-text: Computer-generated image descriptions for blind
  users on a social network service.
\newblock In {\em Proceedings of the 2017 ACM Conference on Computer Supported
  Cooperative Work and Social Computing}, pages 1180--1192. ACM, 2017.

\bibitem{xu2015show}
Kelvin Xu, Jimmy Ba, Ryan Kiros, Kyunghyun Cho, Aaron Courville, Ruslan
  Salakhudinov, Rich Zemel, and Yoshua Bengio.
\newblock Show, attend and tell: Neural image caption generation with visual
  attention.
\newblock In {\em International conference on machine learning}, pages
  2048--2057, 2015.

\bibitem{yang2016review}
Zhilin Yang, Ye Yuan, Yuexin Wu, William~W Cohen, and Ruslan~R Salakhutdinov.
\newblock Review networks for caption generation.
\newblock In {\em Advances in Neural Information Processing Systems}, pages
  2361--2369, 2016.

\bibitem{yao2017incorporating}
Ting Yao, Yingwei Pan, Yehao Li, and Tao Mei.
\newblock Incorporating copying mechanism in image captioning for learning
  novel objects.
\newblock In {\em Proceedings of the IEEE Conference on Computer Vision and
  Pattern Recognition}, pages 6580--6588, 2017.

\bibitem{yao2018exploring}
Ting Yao, Yingwei Pan, Yehao Li, and Tao Mei.
\newblock Exploring visual relationship for image captioning.
\newblock In {\em Proceedings of the European Conference on Computer Vision
  (ECCV)}, pages 684--699, 2018.

\bibitem{yao2017boosting}
Ting Yao, Yingwei Pan, Yehao Li, Zhaofan Qiu, and Tao Mei.
\newblock Boosting image captioning with attributes.
\newblock In {\em Proceedings of the IEEE International Conference on Computer
  Vision}, pages 4894--4902, 2017.

\bibitem{you2016image}
Quanzeng You, Hailin Jin, Zhaowen Wang, Chen Fang, and Jiebo Luo.
\newblock Image captioning with semantic attention.
\newblock In {\em Proceedings of the IEEE conference on computer vision and
  pattern recognition}, pages 4651--4659, 2016.

\bibitem{zhao2017dual}
Wei Zhao, Wei Xu, Min Yang, Jianbo Ye, Zhou Zhao, Yabing Feng, and Yu Qiao.
\newblock Dual learning for cross-domain image captioning.
\newblock In {\em Proceedings of the 2017 ACM on Conference on Information and
  Knowledge Management}, pages 29--38. ACM, 2017.

\bibitem{zhu2017unpaired}
Jun-Yan Zhu, Taesung Park, Phillip Isola, and Alexei~A Efros.
\newblock Unpaired image-to-image translation using cycle-consistent
  adversarial networks.
\newblock In {\em Proceedings of the IEEE International Conference on Computer
  Vision}, pages 2223--2232, 2017.

\end{thebibliography}
}

\end{document}